\documentclass[10pt,twocolumn,letterpaper]{article}

\usepackage{cvpr}
\usepackage{times}
\usepackage{epsfig}
\usepackage{graphicx}
\usepackage{amsmath}
\usepackage{amssymb}
\usepackage{bm}
\usepackage{amsfonts}
\usepackage{threeparttable}
\usepackage{multirow}

\usepackage[breaklinks=true,bookmarks=false]{hyperref}

\cvprfinalcopy 

\def\cvprPaperID{2345} 

\begin{document}

\title{Multi-Person Pose Estimation \\ with Enhanced Channel-wise and Spatial Information}

\author{
Kai Su$^{\dag,1,2}$, Dongdong Yu$^{\dag,2}$, Zhenqi Xu$^{2}$, Xin Geng$^{*,1}$, Changhu Wang$^{*,2}$ \\
$^{1}$School of Computer Science and Engineering, Southeast University, Nanjing, China \\
{\tt\small \{sukai,xgeng\}@seu.edu.cn} \\
$^{2}$ByteDance AI Lab, Beijing, China \\
{\tt\small \{sukai,yudongdong,xuzhenqi,wangchanghu\}@bytedance.com} \\
}

\maketitle
\thispagestyle{empty}

\let\thefootnote\relax\footnotetext{$^{\dag}$ Equal contribution.}
\let\thefootnote\relax\footnotetext{$^{*}$ X. Geng and C. Wang are the corresponding authors.}
\let\thefootnote\relax\footnotetext{This work was performed while Kai Su worked as an intern at ByteDance AI Lab.}
\let\thefootnote\relax\footnotetext{This research was partially supported by the National Key Research \& Development Plan of China (No. 2017YFB1002801), the National Science Foundation of China (61622203), the Collaborative Innovation Center of Novel Software Technology and Industrialization, and the Collaborative Innovation Center of Wireless Communications Technology.}

\begin{abstract}
Multi-person pose estimation is an important but challenging problem in computer vision. Although current approaches have achieved significant progress by fusing the multi-scale feature maps, they pay little attention to enhancing the channel-wise and spatial information of the feature maps. In this paper, we propose two novel modules to perform the enhancement of the information for the multi-person pose estimation. First, a Channel Shuffle Module (CSM) is proposed to adopt the channel shuffle operation on the feature maps with different levels, promoting cross-channel information communication among the pyramid feature maps. Second, a Spatial, Channel-wise Attention Residual Bottleneck (SCARB) is designed to boost the original residual unit with attention mechanism, adaptively highlighting the information of the feature maps both in the spatial and channel-wise context. The effectiveness of our proposed modules is evaluated on the COCO keypoint benchmark, and experimental results show that our approach achieves the state-of-the-art results.
\end{abstract}


\section{Introduction}
Multi-Person Pose Estimation aims to locate body parts for all persons in an image, such as keypoints on the arms, torsos, and the face. It is a fundamental yet challenging task for many computer vision applications like activity recognition \cite{wang2013approach} and human re-identification \cite{zheng2017pose}. Achieving accurate localization results, however, is difficult due to the close-interaction scenarios, occlusions and different human scales.

Recently, due to the involvement of deep convolutional neural networks \cite{lecun1998gradient,he2016deep}, there has been significant progress on the problem of multi-person pose estimation \cite{wei2016convolutional,newell2016stacked,chu2017multi,chen2018cascaded,cao2017realtime,newell2017associative,zanfir2018deep}. Existing approaches for multi-person pose estimation can be roughly classified into two frameworks, i.e., top-down framework \cite{wei2016convolutional,newell2016stacked,chu2017multi,chen2018cascaded} and bottom-up framework \cite{cao2017realtime,newell2017associative,zanfir2018deep}. The former one first detects all human bounding boxes in the image and then estimates the pose within each box independently. The latter one first detects all body keypoints independently and then assembles the detected body joints to form multiple human poses.

\begin{figure}[tb]
	\centering
	\includegraphics[scale=0.30]{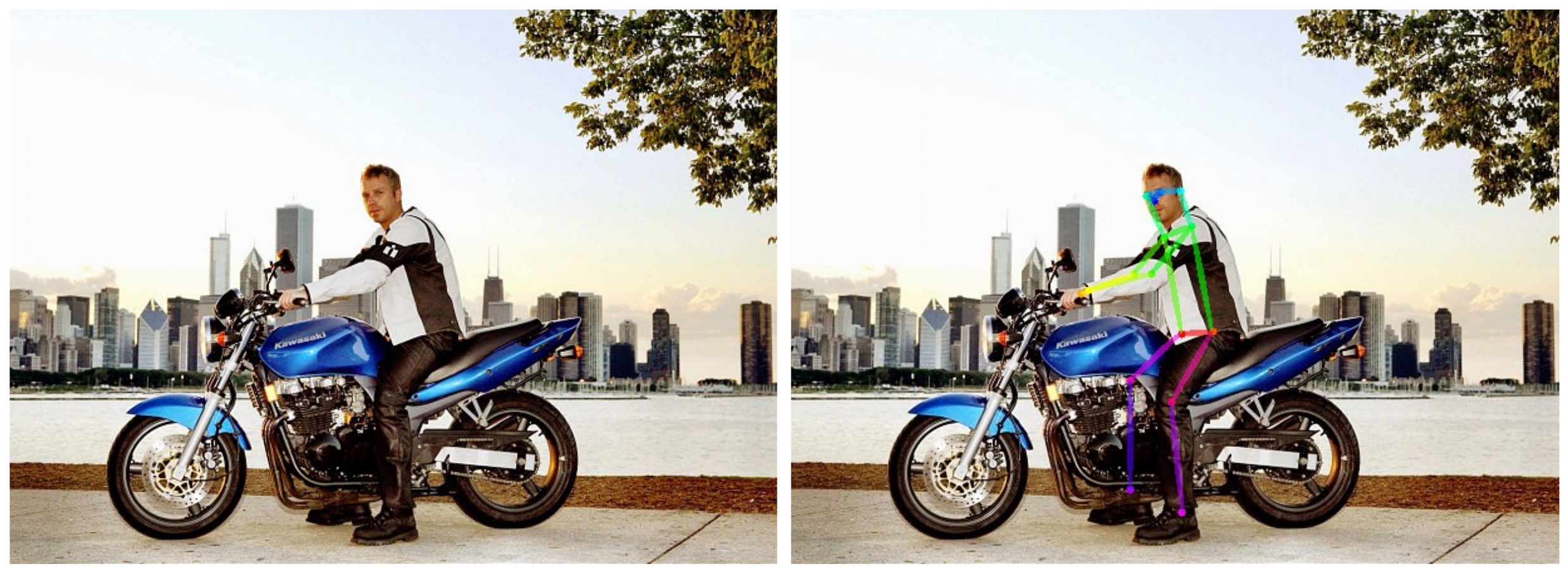}
	\caption{The example of an input image (left) from the COCO test-dev dataset \cite{lin2014microsoft} and its estimated pose (right) from our model.}
	\label{fig:input_image_demo}
\end{figure}

\begin{figure*}[tb]
	\centering
	\includegraphics[scale=0.40]{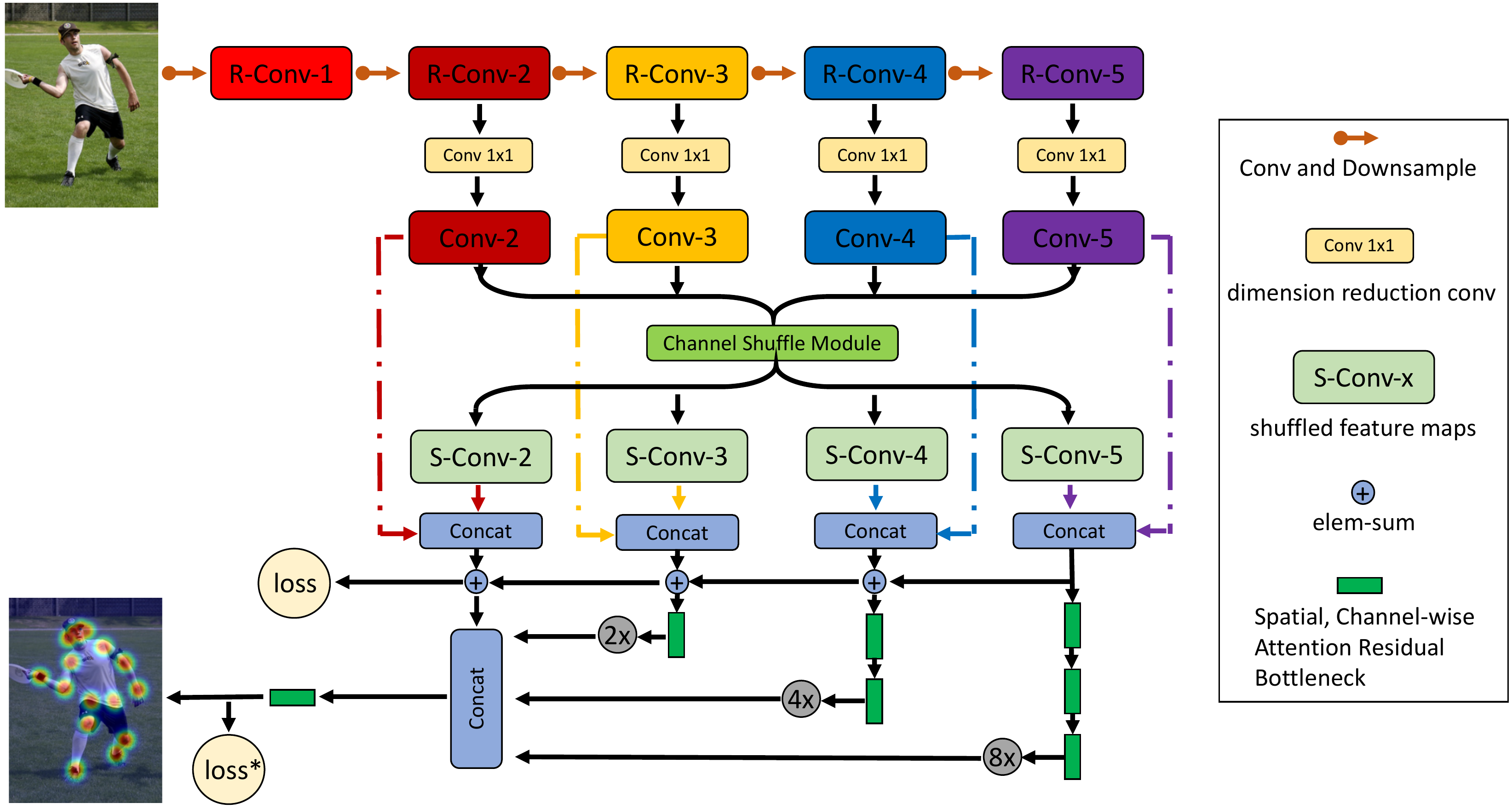}
	\caption{Overview of our architecture. R-Conv-1${\sim}$5 are the last residual blocks of different feature maps from the ResNet backbone \cite{he2016deep}. R-Conv-2${\sim}$5 are first reduced to the same channel dimension of $256$ by $1\times1$ convolution, denoted as Conv-2${\sim}$5. S-Conv-2${\sim}$5 means the corresponding shuffled feature maps after the Channel Shuffle Module. S-Conv-2${\sim}$5 are then concatenated with Conv-2${\sim}$5 as the final enhanced pyramid features. Moreover, a Spatial, Channel-wise Attention Residual Bottleneck is proposed to adaptively enhance the fused pyramid feature responses. Loss denotes the L2 loss and loss* means the L2 loss with Online Hard Keypoints Mining \cite{chen2018cascaded}.}
	\label{fig:architecture}
\end{figure*}

Although great progress has been made, it is still an open problem to achieve accurate localization results. First, on the one hand, high-level feature maps with larger receptive fields are required in some challenging cases to infer the invisible and occluded keypoints, e.g., the right knee of the human in Fig. \ref{fig:input_image_demo}. On the other hand, low-level feature maps with larger resolutions are also helpful to the detailed refinement of the keypoints, e.g., the right ankle of the human in Fig. \ref{fig:input_image_demo}. The trade-off between the low-level and high-level feature maps is more complex in real scenarios. Second, the feature fusion is even dynamic and the fused feature maps always remain redundant. Therefore, the information which is more important to the pose estimation should be adaptively highlighted, e.g., with the help of attention mechanism. According to the above analysis, in this paper, we propose a Channel Shuffle Module (CSM) to further enhance the cross-channel communication between the feature maps across all scales. Moreover, a Spatial, Channel-wise Attention Residual Bottleneck (SCARB) is designed to adaptively enhance the fused feature maps both in the spatial and channel-wise context.

To promote the information communication across the channels among the feature maps at different resolution layers, we further exploit the channel shuffle operation proposed in the ShuffleNet \cite{zhang2018shufflenet}. Different from ShuffleNet, in this paper, we creatively adopt the channel shuffle operation to enable the cross-channel information flow among the feature maps across all scales. To the best of our knowledge, the use of the channel shuffle operation to enhance the information of the feature maps is rarely mentioned in previous work for the multi-person pose estimation. As shown in Fig. \ref{fig:architecture}, the proposed Channel Shuffle Module (CSM) performs on the feature maps Conv-2${\sim}$5 of different resolutions to obtain the shuffled feature maps S-Conv-2${\sim}$5. The idea behind the CSM is that the channel shuffle operation can further recalibrate the interdependencies between the low-level and high-level feature maps.

Moreover, we propose a Spatial, Channel-wise Attention Residual Bottleneck (SCARB), integrating the spatial and channel-wise attention mechanism into the original residual unit \cite{he2016deep}. As shown in Fig. \ref{fig:architecture}, by stacking these SCARBs together, we can adaptively enhance the fused pyramid feature responses both in the spatial and channel-wise context. There is a trend of designing networks with attention mechanism, as it is effective in adaptively highlighting the most informative components of an input feature map. However, spatial and channel-wise attention has little been used in the multi-person pose estimation yet.

As one of the classic methods belonging to the top-down framework, Cascaded Pyramid Network (CPN) \cite{chen2018cascaded} was the winner of the COCO 2017 keypoint Challenge \cite{cocodataset.org}. Since CPN is an effective structure for the multi-person pose estimation, we apply it as the basic network structure in our experiments to investigate the impact of the enhanced channel-wise and spatial information. We evaluate the two proposed modules on the COCO \cite{lin2014microsoft} keypoint benchmark, and ablation studies demonstrate the effectiveness of the Channel Shuffle Module and the Spatial, Channel-wise Attention Residual Bottleneck from various aspects. Experimental results show that our approach achieves the state-of-the-art results.

In summary, our main contributions are three-fold as follows:

\begin{itemize}
\item We propose a Channel Shuffle Module (CSM), which can enhance the cross-channel information communication between the low-level and high-level feature maps.
\item We propose a Spatial, Channel-wise Attention Residual Bottleneck (SCARB), which can adaptively enhance the fused pyramid feature responses both in the spatial and channel-wise context.
\item Our method achieves the state-of-the-art results on the COCO keypoint benchmark.
\end{itemize}

The rest of this paper is organized as follows. First, related work is reviewed. Second, our method is described in details. Then ablation studies are performed to measure the effects of different parts of our system, and the experimental results are reported. Finally, conclusions are given.

\section{Related Work}

This section reviews two aspects related to our method: multi-scale fusion and visual attention mechanism.

\subsection{Multi-scale Fusion Mechanism}

In the previous work for the multi-person pose estimation, large receptive filed is achieved by a sequential architecture in the Convolutional Pose Machines \cite{wei2016convolutional,cao2017realtime} to implicitly capture the long-range spatial relations among multi-parts, producing the increasingly refined estimations. However, low-level information is ignored along the way. Stacked Hourglass Networks \cite{newell2016stacked,newell2017associative} processes the feature maps across all scales to capture various spatial relationships of different resolutions, and adopt the skip layers to preserve spatial information at each resolution. Moreover, the Feature Pyramid Network architecture \cite{lin2017feature} is integrated in the GlobalNet of the Cascaded Pyramid Network \cite{chen2018cascaded}, to maintain both the high-level and low-level information from the feature maps of different scales.

\subsection{Visual Attention Mechanism}

Visual attention has achieved great success in various tasks, such as the network architecture design \cite{hu2018squeeze}, image caption \cite{chen2017sca,you2016image} and pose estimation \cite{chu2017multi}. SE-Net \cite{hu2018squeeze} proposed a ``Squeeze-and-Excitation'' (SE) block to adaptively highlight the channel-wise feature maps by modeling the channel-wise statistics. However, SE block only considers the channel-wise relationship and ignores the importance of the spatial attention in the feature maps. SCA-CNN \cite{chen2017sca} proposed Spatial and Channel-wise Attentions in a CNN for image caption. Spatial and channel-wise attention not only encodes where (i.e., spatial attention) but also introduces what (i.e., channel-wise attention) the important visual attention is in the feature maps. However, spatial and channel-wise attention has little been used in the multi-person pose estimation yet. Chu \textit{et al.} \cite{chu2017multi} proposed the effective multi-context attention model for the human pose estimation. However, our proposed spatial and channel-wise attention residual bottleneck for the multi-person pose estimation has not been mentioned in \cite{chu2017multi} yet.

\section{Method}

An overview of our proposed framework is illustrated in Fig. \ref{fig:architecture}. We adopt the effective Cascaded Pyramid Network (CPN) \cite{chen2018cascaded} as the basic network structure to explore the effects of the Channel Shuffle Module and the Spatial, Channel-wise Attention Residual Bottleneck for the multi-person pose estimation. We first briefly review the structure of the CPN, and then the detailed descriptions of our proposed modules are presented.

\subsection{Revisiting Cascaded Pyramid Network}

Cascaded Pyramid Network (CPN) \cite{chen2018cascaded} is a two-step network structure for the human pose estimation. Given a human box, first, CPN uses the GlobalNet to locate somewhat ``simple'' keypoints based on the FPN architecture \cite{lin2017feature}. Second, CPN adopts the RefineNet with the Online Hard Keypoints Mining mechanism to explicitly address the ``hard'' keypoints.

As shown in Fig. \ref{fig:architecture}, in this paper, for the GlobalNet, the feature maps with different scales (i.e., R-Conv-2${\sim}$5) extracted from the ResNet \cite{he2016deep} backbone are first reduced to the same channel dimension of $256$ by $1\times1$ convolution, denoted as Conv-2${\sim}$5. The proposed Channel Shuffle Module then performs on the Conv-2${\sim}$5 to obtain the shuffled feature maps S-Conv-2${\sim}$5. Finally, S-Conv-2${\sim}$5 are concatenated with the original pyramid features Conv-2${\sim}$5 as the final enhanced pyramid features, which will be used as the U-shape FPN architecture. In addition, for the RefineNet, a boosted residual bottleneck with spatial, channel-wise attention mechanism is proposed to adaptively highlight the feature responses transferred from the GlobalNet both in the spatial and channel-wise context.



\begin{figure}[tb]
	\centering
	\includegraphics[scale=0.40]{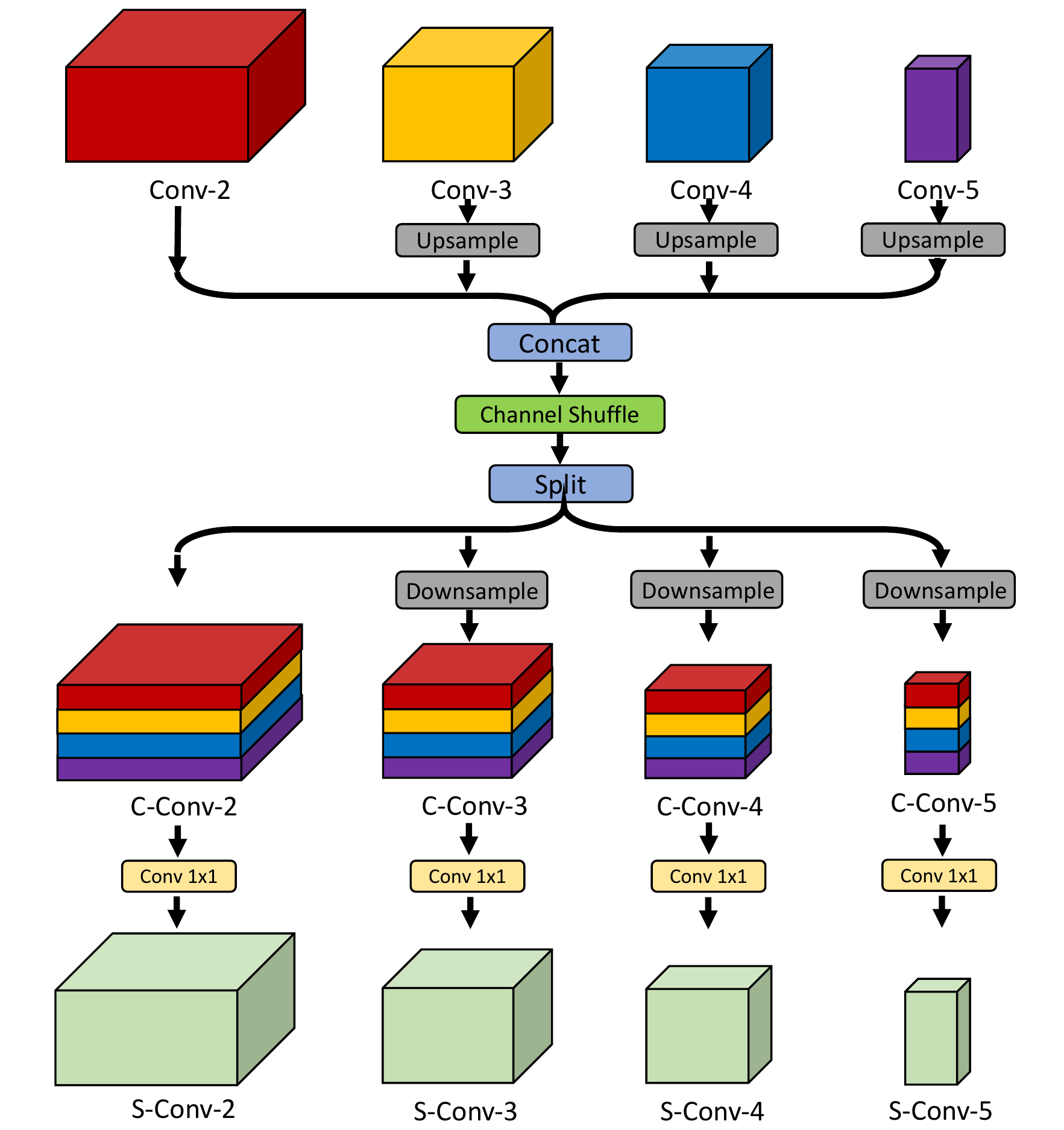}
	\caption{Channel Shuffle Module. The module adopts the channel shuffle operation on the pyramid features Conv-2${\sim}$5 to achieve the shuffled pyramid features S-Conv-2${\sim}$5 with cross-channel communication between different levels. The groups $g$ is set as 4 here.}
	\label{fig:channel_shuffle_module}
\end{figure}

\subsection{CSM: Channel Shuffle Module}

As the levels of the feature maps are greatly enriched by the depth of the layers in the deep convolutional neural networks, many visual tasks have made significant improvements, e.g., image classification \cite{he2016deep}. However, for the multi-person pose estimation, there are still limitations in the trade-off between the low-level and high-level feature maps. The channel information with different characteristics among different levels can complement and reinforce with each other. Motivated by this, we propose the Channel Shuffle Module (CSM) to further recalibrate the interdependencies between the low-level and high-level feature maps.


As shown in Fig. \ref{fig:channel_shuffle_module}, assuming that the pyramid features extracted from the ResNet backbone are denoted as Conv-2${\sim}$5 (with the same channel dimension of $256$). Conv-3${\sim}$5 are first upsampled to the same resolution as the Conv-2, and then these feature maps are concatenated together. After that, the channel shuffle operation is performed on the concatenated features to fuse the complementary channel information among different levels. The shuffled features are then split and downsampled to the original resolution separately, denoted as C-Conv-2${\sim}$5. C-Conv-2${\sim}$5 can be viewed as the features consisting of the complementary channel information from feature maps among different levels. After that, we perform $1 \times 1$ convolution to further fuse C-Conv-2${\sim}$5, and obtain the shuffled features, denoted as S-Conv-2${\sim}$5. We then concatenate the shuffled feature maps S-Conv-2${\sim}$5 with the original pyramid feature maps Conv-2${\sim}$5 to achieve the final enhanced pyramid feature representations. These enhanced pyramid feature maps not only contain the information from the original pyramid features, but also the fused cross-channel information from the shuffled pyramid feature maps.

\subsubsection{Channel Shuffle Operation}

As described in the ShuffleNet \cite{zhang2018shufflenet}, a channel shuffle operation can be modeled as a process composed of ``reshape-transpose-reshape'' operations. Assuming the concatenated features from different levels as $\bm{\Psi}$, and the channel dimension of $\bm{\Psi}$ in this paper is $256*4=1024$. We first reshape the channel dimension of $\bm{\Psi}$ into $(g, c)$, where $g$ is the number of groups, $c = 1024 / g$. Then, we transpose the channel dimension to $(c, g)$, and flatten it back to $1024$. After the channel shuffle operation, $\bm{\Psi}$ is fully related in the channel context. The number of groups $g$ will be discussed in the ablation studies of the experiments.

\subsection{ARB: Attention Residual Bottleneck}

\begin{figure}[tb]
	\centering
	\includegraphics[scale=0.40]{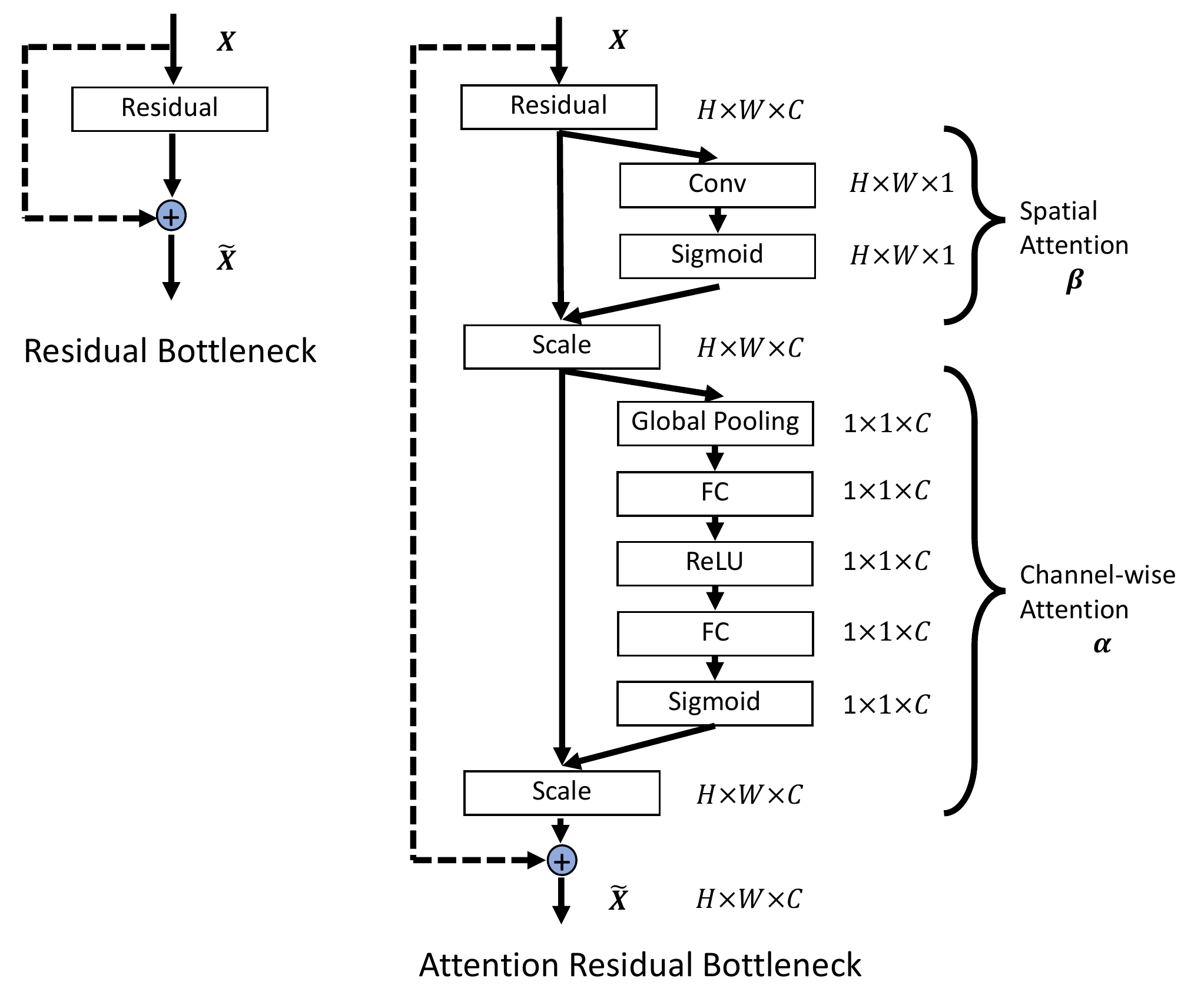}
	\caption{The schema of the original Residual Bottleneck (left) and the Spatial, Channel-wise Attention Residual Bottleneck (right), which is composed of the spatial attention and channel-wise attention. Dashed links indicate the identity mapping.}
	\label{fig:attention_bottleneck}
\end{figure}

Based on the enhanced pyramid feature representations introduced above, we attach our boosted Attention Residual Bottleneck to adaptively enhance the feature responses both in the spatial and channel-wise context. As shown in Fig. \ref{fig:attention_bottleneck}, our Attention Residual Bottleneck learns the spatial attention weights $\bm{\beta}$ and the channel-wise attention weights $\bm{\alpha}$ respectively.

\subsubsection{Spatial Attention}

Applying the whole feature maps may lead to sub-optimal results due to the irrelevant regions. Different from paying attention to the whole image region equally, spatial attention mechanism attempts to adaptively highlight the task-related regions in the feature maps.

Assuming the input of the spatial attention is ${\bm{V}}\in{\mathbb{R}^{H{\times}W{\times}C}}$, and the output of the spatial attention is $\bm{V^{'}}\in{\mathbb{R}^{H{\times}W{\times}C}}$, then we can get $\bm{V^{'}}=\bm{\beta}*{\bm{V}}$, where $*$ means the element-wise multiplication in the spatial context. The spatial-wise attention weights $\bm{\beta}\in{\mathbb{R}^{H{\times}W}}$ is generated by a convolutional operation $\bm{W}\in{\mathbb{R}^{1{\times}1{\times}C}}$ followed by a sigmoid function on the input ${\bm{V}}$, i.e.,

\begin{equation}
\bm{\beta} = Sigmoid(\bm{W}\bm{V}),
\end{equation}
where $\bm{W}$ denotes the convolution weights, and $Sigmoid$ means the sigmoid activation function.

Finally the learned spatial attention weights $\bm{\beta}$ is rescaled on the input $\bm{V}$ to achieve the output $\bm{V^{'}}$.

\begin{equation}
\bm{v^{'}}_{i,j} =  {\beta}_{i,j} * {\bm{v}}_{i,j},
\end{equation}
where $*$ means the element-wise multiplication between the $i,j$-th element of $\bm{\beta}$ and $\bm{V}$ in the spatial context.

\subsubsection{Channel-wise Attention}

Since convolutional filters perform as a pattern detector, and each channel of a feature map after the convolutional operation is the feature activations of the corresponding convolutional filters. The channel-wise attention mechanism can be viewed as a process of adaptively selecting the pattern detectors, which are more important to the task.

Assuming the input of the channel-wise attention is ${\bm{U}}\in{\mathbb{R}^{H{\times}W{\times}C}}$, and the output of the channel-wise attention is ${\bm{U^{'}}}\in{\mathbb{R}^{H{\times}W{\times}C}}$, then we can get $\bm{U^{'}}=\bm{\alpha}*\bm{U}$, where $*$ means the element-wise multiplication in the channel-wise context, $\bm{\alpha}\in{\mathbb{R}^{C}}$ is the channel-wise attention weights. Following the SE-Net \cite{hu2018squeeze}, channel-wise attention can be modeled as a process consisting of two steps, i.e., squeeze and excitation step respectively.

In the squeeze step, global average pooling operation is first performed on the input $\bm{U}$ to generate the channel-wise statistics ${\bm{z}}\in{\mathbb{R}^{C}}$, where the $c$-th element of $\bm{z}$ is calculated by

\begin{equation}
z_c = \frac{1}{H{\times}W} \sum_{i=1}^{H}\sum_{j=1}^{W} \bm{u}_c(i,j),
\end{equation}
where $\bm{u}_c\in{\mathbb{R}^{H{\times}W}}$ is the $c$-th element of the input $\bm{U}$.

In the excitation step, a simple gating mechanism with a sigmoid activation is performed on the channel-wise statistics $\bm{z}$, i.e.,

\begin{equation}
\bm{\alpha} = Sigmoid(\bm{W}_2(\sigma(\bm{W}_1(\bm{z})))),
\end{equation}
where $\bm{W}_1\in{\mathbb{R}^{C\times{C}}}$ and $\bm{W}_2\in{\mathbb{R}^{C\times{C}}}$ denotes two fully connected layers, $\sigma$ means the ReLU activation function \cite{nair2010rectified}, and $Sigmoid$ means the sigmoid activation function.

Finally, the learned channel-wise attention weights $\bm{\alpha}$ is rescaled on the input $\bm{U}$ to achieve the output of the channel-wise attention $\bm{U^{'}}$, i.e.,

\begin{equation}
\bm{u^{'}}_c =  {\alpha}_c * {\bm{u}}_c,
\end{equation}
where $*$ means the element-wise multiplication between the $c$-th element of $\bm{\alpha}$ and $\bm{U}$ in the channel-wise context.

As shown in Fig. \ref{fig:attention_bottleneck}, assuming the input of the residual bottleneck is $\bm{X}\in{\mathbb{R}^{H{\times}W{\times}C}}$, the attention mechanism is performed on the non-identity branch of the residual module, and the spatial, channel-wise attention act before the summation with the identity branch. There are two different implementation orders of the spatial attention and channel-wise attention in the residual bottleneck \cite{he2016deep}, i.e., SCARB: Spatial, Channel-wise Attention Residual Bottleneck and CSARB: Channel-wise, Spatial Attention Residual Bottleneck respectively, which are described as follows.

\subsubsection{SCARB: Spatial, Channel-wise Attention Residual Bottleneck}

The first type applies the spatial attention before the channel-wise attention, as shown in Fig. \ref{fig:attention_bottleneck}. All processes are summarized as follows:

\begin{equation}
\begin{aligned}
&\bm{X^{'}} = F(\bm{X}), \\
&\bm{Y} = \bm\alpha * (\bm\beta*\bm{X^{'}}), \\
&\bm{\widetilde{X}} = \sigma(\bm{X}+\bm{Y}),
\end{aligned}
\end{equation}
where the function $F(\bm{X})$ represents the residual mapping to be learned in the ResNet \cite{he2016deep}, $\bm{\widetilde{X}}$ is the output attention feature maps with the enhanced spatial and channel-wise information.

\subsubsection{CSARB: Channel-wise, Spatial Attention Residual Bottleneck}

Similarly, the second type is a model with channel-wise attention implemented first, i.e.,

\begin{equation}
\begin{aligned}
&\bm{X^{'}} = F(\bm{X}), \\
&\bm{Y} = \bm\beta * (\bm\alpha*\bm{X^{'}}), \\
&\bm{\widetilde{X}} = \sigma(\bm{X}+\bm{Y}).
\end{aligned}
\end{equation}

The choice of the SCARB and CSARB will be discussed in the ablation studies of the experiments.

\section{Experiments}

Our multi-person pose estimation system follows the top-down pipeline. First, a human detector is applied to generate all human bounding boxes in the image. Then for each human bounding box, we apply our proposed network to predict the corresponding human pose.

\subsection{Experimental Setup}

\subsubsection{Datasets and Evaluation Criterion}

We evaluate our model on the challenging COCO keypoint benchmark \cite{lin2014microsoft}. Our models are only trained on the COCO trainval dataset (includes $57K$ images and $150K$ person instances) with no extra data involved. Ablation studies are validated on the COCO minival dataset (includes $5K$ images). The final results are reported on the COCO test-dev dataset (includes $20K$ images) compared with the public state-of-the-art results. We use the official evaluation metric \cite{lin2014microsoft} that reports the OKS-based AP (average precision) in the experiments, where the OKS (object keypoints similarity) defines the similarity between the predicted pose and the ground truth pose.

\subsubsection{Training Details}

Our pose estimation model is implemented in Pytorch \cite{paszke2017automatic}. For the training, $4$ V$100$ GPUs on a server are used. Adam \cite{kingma2014adam} optimizer is adpoted. The base learning rate is set to $5e-4$, and is decreased by a factor of $0.1$ at $90$ and $120$ epochs, and finally we train for $140$ epochs. The input size of the image for the network is made to a fixed aspect ratio of height : width = $4 : 3$, e.g., $256\times192$ is used as the default resolution, the same as the CPN \cite{chen2018cascaded}. L2 loss is used for the GlobalNet, and following the CPN, we only punish the top $8$ keypoints losses in the Online Hard Keypoint Mining of the RefineNet. Data augmentation during the training includes the random rotation ($-40^\circ \sim +40^\circ$) and the random scale ($0.7 \sim 1.3$).

Our ResNet backbone is initialized with the weights of the public-released Imagenet \cite{russakovsky2015imagenet} pre-trained model. ResNet backbones with $50$, $101$ and $152$ layers are experimented. ResNet-$50$ is used by default, unless otherwise noted.

\newcommand{\tabincell}[2]{
	\begin{tabular}{@{}#1@{}}#2\end{tabular}
}
\begin{table}[tb]
	\centering
	\caption{Ablation study on the Channel Shuffle Module (CSM) with different groups $g$ on the COCO minival dataset. CSM-$g$ denotes the Channel Shuffle Module with $g$ groups. The Attention Residual Bottleneck is not used in this experiment.}\label{table:groupsMinivalPerformance}
	
	\resizebox{0.43\columnwidth}{!}{
	\begin{tabular}{cc}
		\hline
		Method & \tabincell{c}{AP} \\
		\hline
		CPN (baseline) & $69.4$ \\
		CPN + CSM-$2$ & $70.4$ \\
		CPN + CSM-$4$ & $\textbf{71.7}$ \\
		CPN + CSM-$8$ & $71.4$ \\
		CPN + CSM-$16$ & $71.2$ \\
		CPN + CSM-$32$ & $70.1$ \\
		CPN + CSM-$64$ & $70.7$ \\
		CPN + CSM-$128$ & $71.0$ \\
		CPN + CSM-$256$ & $71.6$ \\
		\hline
	\end{tabular}
	}
\end{table}

\begin{table}[tb]
	\centering
	\caption{Ablation study on the Attention Residual Bottleneck on the COCO minival dataset. SCARB denotes the Spatial, Channel-wise Attention Residual Bottleneck, CSARB denotes the Channel-wise, Spatial Attention Residual Bottleneck. The Channel Shuffle Module is not used in this experiment.}\label{table:attentionMinivalPerformance}
	
	\resizebox{0.40\columnwidth}{!}{
	\begin{tabular}{cc}
		\hline
		Method & \tabincell{c}{AP} \\
		\hline
		CPN (baseline) & $69.4$ \\
		CPN + CSARB & $70.4$ \\
		CPN + SCARB & $\textbf{70.8}$ \\
		\hline
	\end{tabular}
	}
\end{table}

\begin{table}[tb]
	\centering
	\caption{Component analysis on the Channel Shuffle Module with $4$ groups (CSM-$4$) and the Spatial, Channel-wise Attention Residual Bottleneck (SCARB) on the COCO minival dataset. Based on the baseline CPN \cite{chen2018cascaded}, we gradually add the CSM-$4$ and SCARB for ablation studies. The last line shows the total improvement compared with the baseline CPN.}\label{table:componentMinivalPerformance}
	
	\resizebox{0.90\columnwidth}{!}{
	\begin{tabular}{cccc}
		\hline
		Method & CSM-$4$ & SCARB & \tabincell{c}{AP} \\
		\hline
		CPN (baseline) &  &  & $69.4$ \\
		CPN + CSM-$4$ & $\surd$ &  & $71.7$ \\
		CPN + SCARB &  & $\surd$ & $70.8$ \\
		CPN + CSM-$4$ + SCARB & $\surd$ & $\surd$ & $\textbf{72.1}$ \\
		\hline
	\end{tabular}
	}
\end{table}

\subsubsection{Testing Details}

For the testing, a top-down pipeline is applied. For the COCO minival dataset, we use the human detection results provided by the CPN \cite{chen2018cascaded} for the fair comparison, which reports the human detection AP $55.3$. For the COCO test-dev dataset, we adopt the SNIPER \cite{singh2018sniper} as the human detector, which achieves the human detection AP $58.1$. Following the common practice in \cite{chen2018cascaded,newell2016stacked}, the keypoints are estimated on the averaged heatmaps of the original and flipped image. A quarter offset in the direction from the highest response to the second highest response is used to obtain the final keypoints.

\subsection{Component Ablation Studies}

In this section, we conduct the ablation studies on the Channel Shuffle Module and the Attention Residual Bottleneck on the COCO minival dataset. The ResNet-$50$ backbone and the input size of $256\times192$ are used by default in the all ablation studies.

\begin{table}[tb]
	\centering
	\caption{Comparison with the $8$-stage Hourglass \cite{newell2016stacked}, CPN \cite{chen2018cascaded} and Simple Baselines \cite{xiao2018simple} on the COCO minival dataset. Their results are cited from \cite{chen2018cascaded, xiao2018simple}. ``*'' means the model training with the Online Hard Keypoints Mining.}\label{table:cocoMinivalPerformance}
	
	\resizebox{0.90\columnwidth}{!}{
	\begin{tabular}{cccc}
		\hline
		Method & \tabincell{c}{Backbone} & \tabincell{c}{Input Size} & \tabincell{c}{AP} \\
		\hline
		$8$-stage Hourglass & - & $256\times192$ & $66.9$ \\
		$8$-stage Hourglass & - & $256\times256$ & $67.1$ \\
		CPN (baseline) & ResNet-$50$ & $256\times192$ & $68.6$ \\ 
		CPN (baseline) & ResNet-$50$ & $384\times288$ & $70.6$ \\ 
		CPN* (baseline) & ResNet-$50$ & $256\times192$ & $69.4$ \\ 
		CPN* (baseline) & ResNet-$50$ & $384\times288$ & $71.6$ \\
		Simple Baselines & ResNet-$50$ & $256\times192$ & $70.6$ \\
		Simple Baselines & ResNet-$50$ & $384\times288$ & $72.2$ \\
		\hline
		Ours* & ResNet-$50$ & $256\times192$ & $\textbf{72.1}$ \\
		Ours* & ResNet-$50$ & $384\times288$ & $\textbf{73.8}$ \\
		\hline
	\end{tabular}
	}
\end{table}

\begin{figure}[tb]
	\centering
	\includegraphics[scale=0.30]{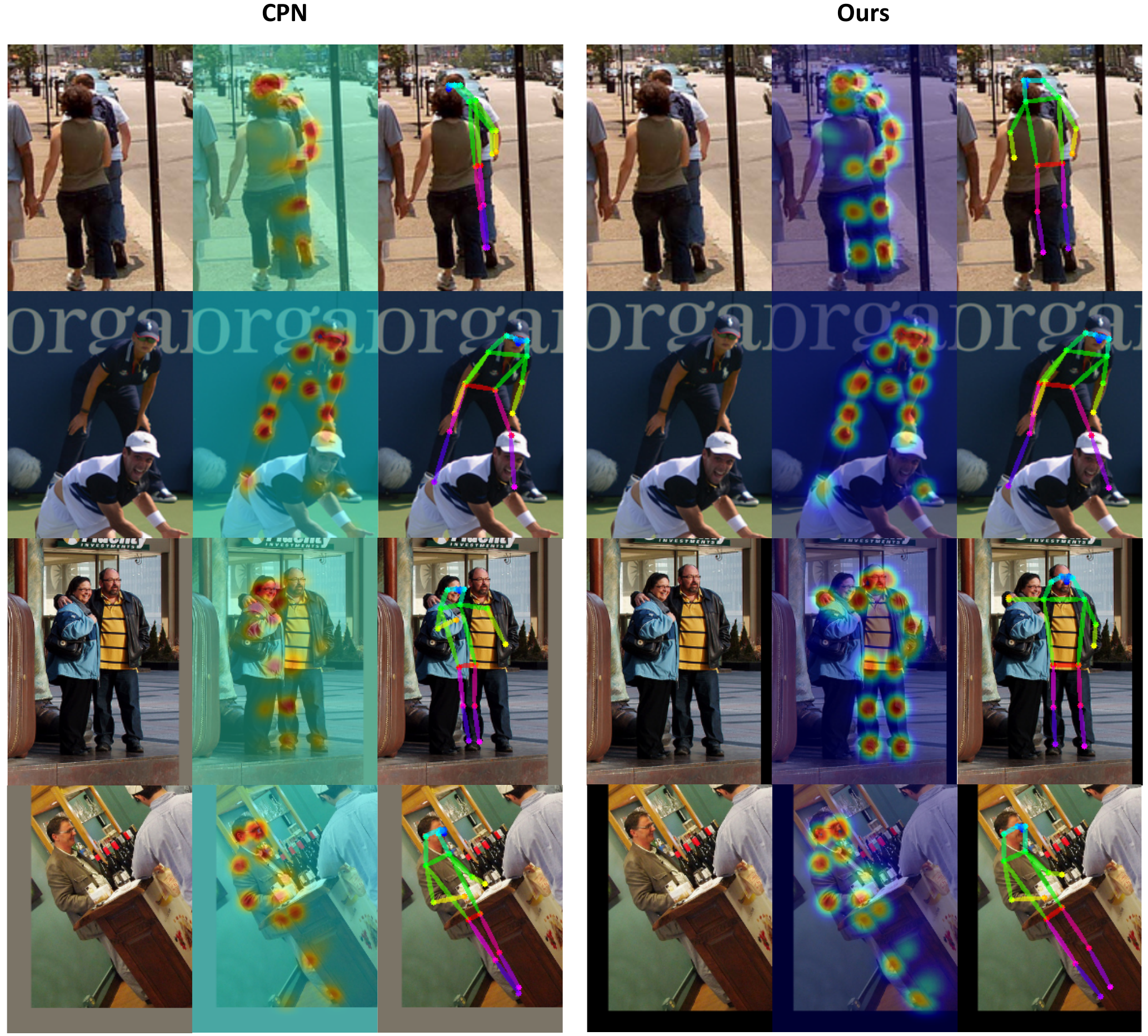}
	\caption{Visual heatmaps of CPN and our model on the COCO minival dataset. From left to right are input images, predicted heatmaps and predicted poses. Best viewed in zoom and color.}
	\label{fig:heatmaps}
\end{figure}

\begin{table*}[tb]
	\centering
	\caption{Comparison of final results on the COCO test-dev dataset. \textbf{Top:} methods in the literature, trained only with the COCO trainval dataset. \textbf{Middle:} results submitted to the COCO test-dev leaderboard \cite{cocodataset.org}. ``*'' means that the method involves extra data for the training. ``+'' indicates the results using the ensembled models. \textbf{Bottom:} the results of our single model, trained only with the COCO trainval dataset. $\divideontimes$ indicates the results using the single model with flip and rotation testing strategy.}\label{table:testdevPerformance}
	
	\resizebox{1.64\columnwidth}{!}{
	\begin{tabular}{ccccccccc}
		\hline
		Method & \tabincell{c}{Backbone} & \tabincell{c}{Input Size} & \tabincell{c}{AP} & \tabincell{c}{AP $.5$} & \tabincell{c}{AP $.75$} & \tabincell{c}{AP (M)} & \tabincell{c}{AP (L)} & \tabincell{c}{AR} \\
		\hline
		CMU-Pose \cite{cao2017realtime} & - & - & $61.8$ & $84.9$ & $67.5$ & $57.1$ & $68.2$ & $66.5$ \\
		Mask-RCNN \cite{he2017mask} & ResNet-$50$-FPN & - & $63.1$ & $87.3$ & $68.7$ & $57.8$ & $71.4$ & - \\
		\tabincell{c}{Associative Embedding \\ \cite{newell2017associative}} & - & $512\times512$ & $65.5$ & $86.8$ & $72.3$ & $60.6$ & $72.6$ & $70.2$ \\
		G-RMI \cite{papandreou2017towards}  & ResNet-$101$ & $353\times257$ & $64.9$ & $85.5$ & $71.3$ & $62.3$ & $70.0$ & $69.7$ \\
		CPN \cite{chen2018cascaded} & ResNet-Inception & $384\times288$ & $72.1$ & $91.4$ & $80.0$ & $68.7$ & $77.2$ & $78.5$ \\
		Simple Baselines \cite{xiao2018simple} & ResNet-$101$ & $384\times288$ & $73.2$ & $91.4$ & $80.9$ & $69.7$ & $79.5$ & $78.6$ \\
		Simple Baselines \cite{xiao2018simple} & ResNet-$152$ & $384\times288$ & $73.8$ & $91.7$ & $81.2$ & $70.3$ & $80.0$ & $79.1$ \\
		\hline
		\tabincell{c}{FAIR Mask R-CNN* \\ \cite{cocodataset.org}} & ResNet-$101$-FPN & - & $69.2$ & $90.4$ & $77.0$ & $64.9$ & $76.3$ & $75.2$ \\
		G-RMI* \cite{cocodataset.org} & ResNet-$152$ & $353\times257$ & $71.0$ & $87.9$ & $77.7$ & $69.0$ & $75.2$ & $75.8$ \\
		oks* \cite{cocodataset.org} & - & - & $72.0$ & $90.3$ & $79.7$ & $67.6$ & $78.4$ & $77.1$ \\
		bangbangren*+ \cite{cocodataset.org} & ResNet-$101$ & - & $72.8$ & $89.4$ & $79.6$ & $68.6$ & $80.0$ & $78.7$ \\
		CPN+ \cite{cocodataset.org} & ResNet-Inception & $384\times288$ & $73.0$ & $91.7$ & $80.9$ & $69.5$ & $78.1$ & $79.0$ \\
		\hline
		Ours & ResNet-$50$ & $256\times192$ & $71.4$ & $91.3$ & $79.8$ & $68.3$ & $77.1$ & $77.1$ \\
		Ours & ResNet-$50$ & $384\times288$ & $73.2$ & $91.9$ & $81.0$ & $69.6$ & $79.3$ & $78.5$ \\
		Ours & ResNet-$101$ & $256\times192$ & $71.8$ & $91.3$ & $80.1$ & $68.7$ & $77.3$ & $78.8$ \\
		Ours & ResNet-$101$ & $384\times288$ & $73.8$ & $91.7$ & $81.4$ & $70.4$ & $79.6$ & $80.3$ \\
		Ours & ResNet-$152$ & $256\times192$ & $72.3$ & $91.4$ & $80.6$ & $69.2$ & $77.8$ & $79.2$ \\
		Ours & ResNet-$152$ & $384\times288$ & $74.3$ & $91.8$ & $81.9$ & $70.7$ & $80.2$ & $80.5$ \\
		Ours$\divideontimes$ & ResNet-$101$ & $384\times288$ & $74.1$ & $91.8$ & $81.7$ & $70.6$ & $80.0$ & $80.4$ \\
		Ours$\divideontimes$ & ResNet-$152$ & $384\times288$ & $\textbf{74.6}$ & $\textbf{91.8}$ & $\textbf{82.1}$ & $\textbf{70.9}$ & $\textbf{80.6}$ & $\textbf{80.7}$ \\
		\hline
	\end{tabular}
	}
\end{table*}

\subsubsection{Groups $g$ in the Channel Shuffle Module}

In this experiment, we explore the performances of the Channel Shuffle Module with different groups on the COCO minival dataset. CSM-$g$ denotes the Channel Shuffle Module with $g$ groups and the groups $g$ controls the degree of the cross-channel feature maps fusion. The ResNet-$50$ backbone and the input size of $256\times192$ are used by default, and the Attention Residual Bottleneck is not used here. As the Table \ref{table:groupsMinivalPerformance} shows, $4$ groups achieves the best AP of $71.7$. It indicates that when only using the Channel Shuffle Module with $4$ groups (CSM-$4$), we can achieve $2.3$ AP improvement compared with the baseline CPN. Therefore, $4$ groups (CSM-$4$) is selected finally.

\subsubsection{Attention Residual Bottleneck: SCARB and CSARB}

\begin{figure*}[tb]
	\centering
	\includegraphics[scale=0.4]{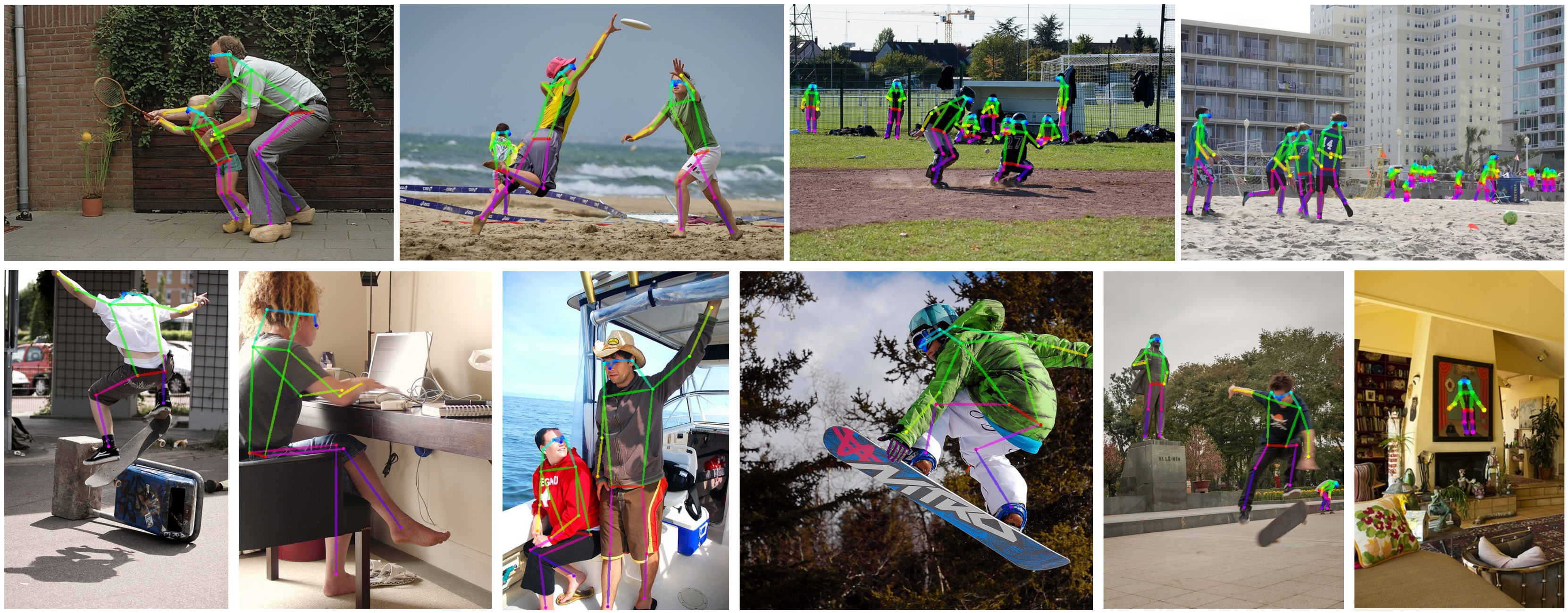}
	\caption{Qualitative results of our model on the COCO test-dev dataset. Our model deals well with the diverse poses, occlusions and cluttered scenes.}
	\label{fig:test_dev_demo}
\end{figure*}

In this experiment, we explore the effects of different implementation orders of the spatial attention and the channel-wise attention in the Attention Residual Bottleneck, i.e., SCARB and CSARB. The ResNet-$50$ backbone and the input size of $256\times192$ are used by default, and the Channel Shuffle Module is not used here. As shown in Table \ref{table:attentionMinivalPerformance}, the SCARB achieves the best AP of $70.8$. It indicates that when only using the SCARB, our model outperforms the baseline CPN by $1.4$ AP. Therefore, SCARB is selected by default.

\subsubsection{Component Analysis}

In this experiment, we analyze the importance of each proposed component on the COCO minival dataset, i.e., the Channel Shuffle Module and the Attention Residual Bottleneck. According to Table \ref{table:groupsMinivalPerformance} and \ref{table:attentionMinivalPerformance}, the Channel Shuffle Module with $4$ groups (CSM-$4$) and the Spatial, Channel-wise Attention Residual Bottleneck (SCARB) are selected finally. Accroding to Table \ref{table:componentMinivalPerformance}, compared with the $69.4$ AP of the baseline CPN, with only the CSM-$4$ used, we can achieve $71.7$ AP, and with only the SCARB used, we can achieve $70.8$ AP. With all the proposed components used, we can achieve $72.1$ AP,  with the improvement of $2.7$ AP over the baseline CPN.

\subsection{Comparisons on COCO minival dataset}

Table \ref{table:cocoMinivalPerformance} compares our model with the $8$-stage Hourglass \cite{newell2016stacked}, CPN \cite{chen2018cascaded} and Simple Baselines \cite{xiao2018simple} on the COCO minival dataset. The human detection AP of the $8$-stage Hourglass and CPN are the same $55.3$ as ours. The human detection AP reported in the Simple Baselines is $56.4$. Compared with the $8$-stage Hourglass, both methods use an input size of $256\times192$, our model has an improvement of $5.2$ AP. CPN and our model both use the Online Hard Keypoints Mining, our model outperforms the CPN by $2.7$ AP for the input size of $256\times192$ and $2.2$ AP for the input size of $384\times288$. Compared with the Simple Baselines, our model outperforms $1.5$ AP for the input size of $256\times192$, and $1.6$ AP for the input size of $384\times288$. Fig. \ref{fig:test_dev_demo} demonstrates the visual heatmaps of CPN and our model on the COCO minival dataset. As shown in Fig. \ref{fig:test_dev_demo}, our model still works in the scenarios (e.g., close-interactions, occlusions) where CPN can not well deal with.

\subsection{Experiments on COCO test-dev dataset}

In this section, we compare our model with the state-of-the-art methods on the COCO test-dev dataset, and analyze the relationships between the human detection performance and the corresponding pose estimation performance.

\subsubsection{Comparison with the state-of-the-art Methods}

Table \ref{table:testdevPerformance} compares our model with other state-of-the-art methods on the COCO test-dev dataset. For the CPN, a human detector with human detection AP $62.9$ on the COCO minival dataset is used. For the Simple Baselines, a human detector with human detection AP $60.9$ on the COCO test-dev dataset is used. Without extra data for training, our single model can achieve $73.8$ AP with the ResNet-$101$ backbone, and $74.3$ AP with the ResNet-$152$ backbone, which outperform both CPN's single model $72.1$ AP, ensembled model $73.0$ AP and Simple Baselines $73.8$ AP. Moreover, when using the averaged heatmaps of the original, flipped and rotated ($+30^\circ$, $-30^\circ$ is used here) images, our single model can achieve $74.1$ AP with the ResNet-$101$ backbone, and $74.6$ AP with the ResNet-$152$ backbone. Fig. \ref{fig:test_dev_demo} demonstrates the poses predicted by our model on the COCO test-dev dataset.

\begin{table}[tb]
	\centering
	\caption{Comparison between the human detection performance and the pose estimation performance on the COCO test-dev dataset. All pose estimation methods are trained with the ResNet-$152$ backbone and the $384\times288$ input size.}\label{table:testdevDetectionPerformance}
	
	\resizebox{1.0\columnwidth}{!}{
	\begin{tabular}{cccc}
		\hline
		Pose Method & Det Method & \tabincell{c}{Human AP} & \tabincell{c}{Pose AP} \\
		\hline
		Simple Baselines \cite{xiao2018simple} & Faster-RCNN \cite{ren2015faster} & $\textbf{60.9}$ & $73.8$ \\
		Ours & Deformable \cite{dai2017deformable} & $45.8$ & $72.9$ \\
		Ours & - \cite{chen2018cascaded} & $57.2$ & $73.8$ \\
		Ours & SNIPER \cite{singh2018sniper} & $58.1$ & $\textbf{74.3}$ \\
		\hline
	\end{tabular}
	}
\end{table}

\subsubsection{Human Detection Performance}

Table \ref{table:testdevDetectionPerformance} shows the relationships between the human detection performance and the corresponding pose estimation performance on the COCO test-dev dataset. Our model and Simple Baselines \cite{xiao2018simple} are compared in this experiment. Both models are trained with the ResNet-$152$ backbone and the $384\times288$ input size. The Simple Baselines adopts the Faster-RCNN \cite{ren2015faster} as the human detector, which reports the human detection AP $60.9$ in their paper. For our model, we adopt the SNIPER \cite{singh2018sniper} as the human detector, which achieves the human detection AP $58.1$. Moreover, we also use the Deformable Convolutional Networks \cite{dai2017deformable} (achieves the human detection AP $45.8$) and the human detection results provided by the CPN \cite{chen2018cascaded} (reports the human detection AP $57.2$) for comparison.

From the table, we can see that the pose estimation AP gains increasingly when the human detection AP increases. For example, when the human detection AP increases from $57.2$ to $58.1$, the pose estimation AP of our model increases from $73.8$ to $74.3$. However, although the human detection AP $60.9$ of the Simple Baselines is higher than ours $58.1$ AP, the pose estimation AP $73.8$ of the Simple Baselines is lower than ours $74.3$ AP. Therefore, we can conclude that it is more important to enhance the accuracy of the pose estimator than the human detector.



\section{Conclusions}

In this paper, we tackle the multi-person pose estimation with the top-down pipeline. The Channel Shuffle Module (CSM) is proposed to promote the cross-channel information communication among the feature maps across all scales, and a Spatial, Channel-wise Attention Residual Bottleneck (SCARB) is designed to adaptively highlight the fused pyramid feature maps both in the spatial and channel-wise context. Overall, our model achieves the state-of-the-art performance on the COCO keypoint benchmark.


{\small
\bibliographystyle{ieee_fullname}
\bibliography{egbib}
}

\end{document}